\documentclass[letterpaper, 10 pt, conference]{ieeeconf}  

\IEEEoverridecommandlockouts                              

\title{\LARGE \bf
Learning Based MPC for Autonomous Driving Using a Low Dimensional Residual Model
}

\author{Yaoyu Li, Chaosheng Huang$^{\ast}$, Dongsheng Yang, Wenbo Liu and Jun Li
\thanks{*Research supported by the National Science Foundation of China Project: 20211300183. (Corresponding author: Chaosheng Huang)}
\thanks{Yaoyu Li, Chaosheng Huang, Dongsheng Yang, Wenbo Liu and Jun Li are with School of Vehicle and Mobility, Tsinghua University, Beijing 100084, China. (e-mail: liyy20@mails.tsinghua.edu.cn; huangchaosheng@tsinghua.edu.cn; 13823740237@163.com; lwb20011108@163.com; lijun1958@tsinghua.edu.cn)}%
}

\usepackage{amsmath}

\usepackage{float}
\usepackage{booktabs}
\usepackage{multirow}

\usepackage{graphicx}
\usepackage{subcaption}
\usepackage{stfloats}

\begin{document}

\maketitle
\thispagestyle{empty}
\pagestyle{empty}

\begin{abstract}
In this paper, a learning based Model Predictive Control (MPC) using a low dimensional residual model is proposed for autonomous driving. One of the critical challenge in autonomous driving is the complexity of vehicle dynamics, which impedes the formulation of accurate vehicle model. Inaccurate vehicle model can significantly impact the performance of MPC controller. To address this issue, this paper decomposes the nominal vehicle model into invariable and variable elements. The accuracy of invariable component is ensured by calibration, while the deviations in the variable elements are learned by a low-dimensional residual model. The features of residual model are selected as the physical variables most correlated with nominal model errors. Physical constraints among these features are formulated to explicitly define the valid region within the feature space. The formulated model and constraints are incorporated into the MPC framework and validated through both simulation and real vehicle experiments. The results indicate that the proposed method significantly enhances the model accuracy and controller performance. 

\end{abstract}

\begin{keywords}
Model Learning, Model Predictive Control, Vehicle Dynamics, Autonomous Driving.
\end{keywords}

\section{INTRODUCTION}
The continuous expansion of operational domain for autonomous vehicles has directly driven advancements in autonomous driving motion control \cite{2021av1}. Due to the capacity of handling complex constraints and generating optimal control inputs, Model Predictive Control (MPC) has been widely employed in autonomous driving motion control \cite{stano2023model}. In MPC, the accuracy of the vehicle model is critical, as it directly determines the performance of controller \cite{chen2020implementation}. An accurate vehicle model is therefore essential for MPC.

Due to the multiple complex subsystems in vehicles and the mathematically elusive interactions between tires and the road, deriving an accurate mathematical representation of vehicle dynamics has long been a core challenge for autonomous driving control \cite{yang2013overview,schramm2014vehicle}. Over the past few decades, numerous MPC controllers that employ physical models to represent vehicle dynamics have been proposed. The models employed in MPC have evolved from simple kinematic models \cite{de2005feedback, gao2021robust} to more detailed single-track models \cite{hu2015output,brown2019coordinating} that incorporate tire characteristics \cite{christ2021time}. The physical models in MPC must accurately capture vehicle dynamics while maintaining limited complexity to enable real-time optimization, leading to suboptimal
solutions. Furthermore, vehicle dynamics can vary during operation due to factors such as temperature fluctuations and tire wear, rendering pre-defined physical models insufficiently adaptive to these dynamics changes \cite{spielberg2021neural}.

In recent years, data-driven algorithms have received significant attention in the academic community due to their excellent ability to approximate unknown functions \cite{cully2015robots,hewing2020learning}. 
These algorithms have increasingly been applied to the modeling of vehicle dynamics \cite{spielberg2021neural,kuutti2020survey,spielberg2019neural}.
Especially, Gaussian Process (GP) based residual model has demonstrated the capability to extract unknown dynamics from sensor data, thereby compensating for the deviations in nominal model \cite{mesbah2022fusion,ning2023scalable}. 
In \cite{ostafew2014learning,ostafew2016robust} and \cite{mckinnon2017learning}, GPs have been employed to enhance the kinematic nominal model of a low-speed vehicle.
To accommodate a wider range of operational domain, MPC based on single-track nominal model is implemented for high-speed autonomous racing tasks in \cite{hewing2018cautious,hewing2019cautious} and \cite{kabzan2020amz}, while GP-based residual model is employed to compensate for dynamics errors in the nominal model using dynamics states as features.

The predictive accuracy of GP is intrinsically dependent on the data within the training set \cite{williams2006gaussian}. However, due to the complexity of vehicle dynamics, the residual model associated with the single-track dynamics model typically involves high dimensional features, necessitating an extensive training set to cover the entire feature space. In practice, it is infeasible to construct the training set by driving a vehicle to traverse every region of the high dimensional feature space. Moreover, the computational burden imposed by such a large training set is prohibitive \cite{kabzan2019learning}. High dimensional features severely constrains the residual model and controller performance, limiting their effectiveness to a narrow range of operational scenarios. The feature for residual model requires further investigation to reduce the dimensionality.

\begin{figure}[!t]
\centering
\includegraphics[width=1.0\linewidth]{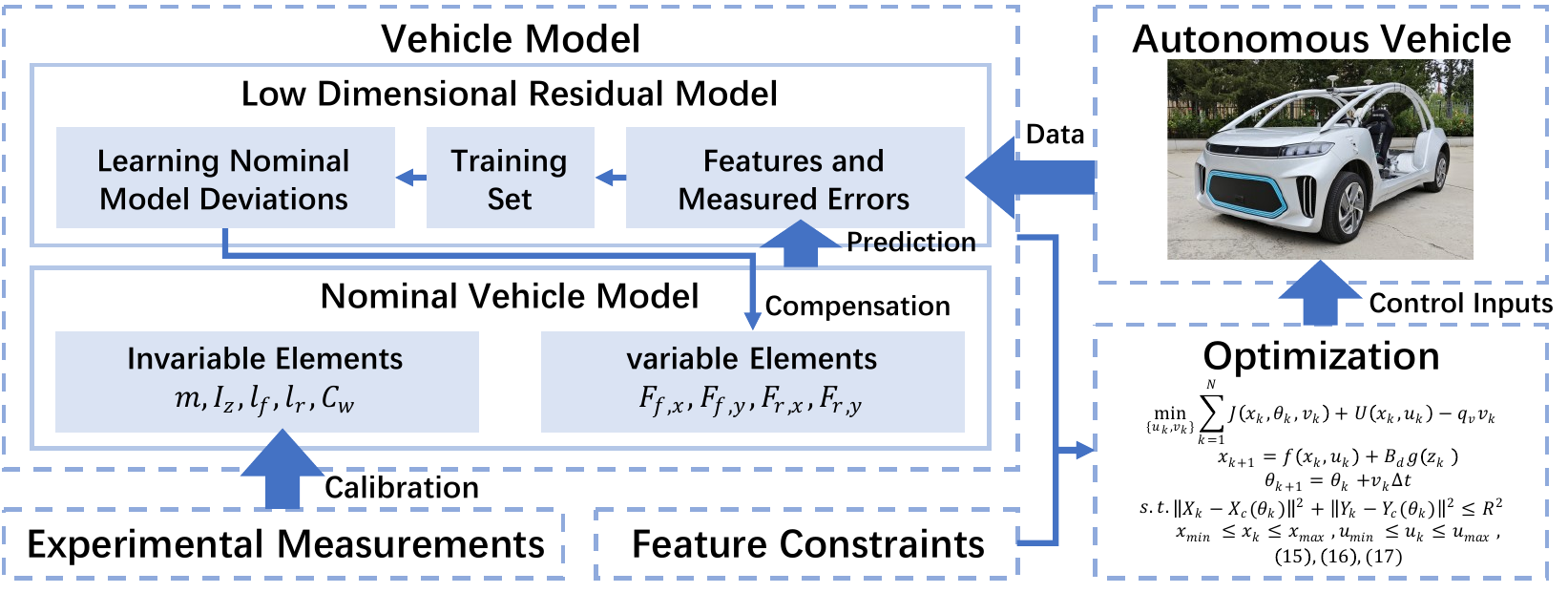}
\caption{The learning based MPC framework.}
\label{MPC_Framework}
\vspace{-5.0ex}
\end{figure}

To address these challenges, this paper proposes a learning based MPC controller using a low dimensional residual model for autonomous driving as shown in Fig. \ref{MPC_Framework}. The vehicle model in the MPC comprises a single-track nominal dynamics model and a low dimensional residual model. Constraints among features of residual model are designed and integrated into the controller. The main contributions are as follows:

\begin{itemize}

\item The nominal vehicle model is decomposed into invariable and variable elements. The accuracy of the invariable elements is guaranteed through experimental measurements, while deviations in the variable elements are learned by a low dimensional residual model. Physical variables most correlated with nominal model deviations are selected as features of residual model. This decomposition reduces the dimensionality of the residual model features. 
\item Physical constraints among features of residual model are formulated to explicitly define the valid region within feature space and ensure the feasibility of proposed model and controller.
\item The proposed learning based MPC controller is formulated and validated through simulations and real vehicle experiments, demonstrating its effectiveness in improving model accuracy and control performance.

\end{itemize}

The rest of the paper is organized as follows: Section \ref{section: The Vehicle Model} introduces the proposed vehicle model. Section \ref{learning based mpc formulation} outlines the MPC formulation. In Section \ref{results}, the simulation and real-world experimental results are presented. Finally, Section \ref{conclusions} concludes with the main point of the proposed method.

\section{THE VEHICLE MODEL}
\label{section: The Vehicle Model}
\begin{figure}[!t]
\centering
\vspace{1.0ex}
\includegraphics[width=0.8\linewidth]{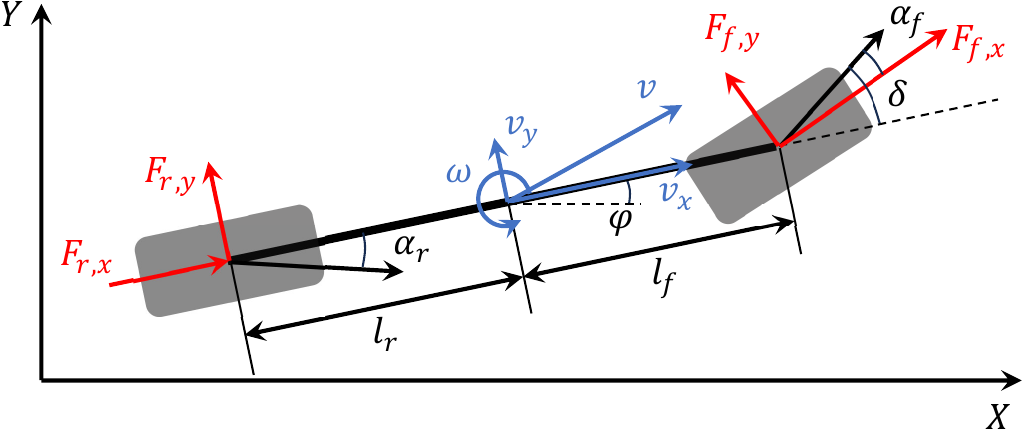}
\vspace{-1.0ex}
\caption{Vehicle dynamics model.}
\label{3dof}
\vspace{-4.0ex}
\end{figure}

In this section, the vehicle model shown in Fig. \ref{3dof} is introduced. The states of the model are:
$$
\mathbf{x}=[X,Y,\varphi,v_x,v_y,\omega,T,\delta]^T. 
$$
The states include the global position $X$ and $Y$, yaw angle $\varphi$, vehicle longitudinal velocity $v_x$, vehicle lateral velocity $v_y$, yaw rate $\omega$, command torque $T$ and steering angle $\delta$. The control inputs are the change of command torque $\Delta T$ and steering angle $\Delta \delta$:
$$
\mathbf{u}=[\Delta T, \Delta \delta]^T.
$$

The vehicle model proposed in this paper composed of a nominal model and a GP-based residual model. The nominal model captures the essential vehicle dynamics, while the residual model compensates for the discrepancies between the nominal model and the actual vehicle dynamics. The system dynamics in discrete time is formulated as:
\begin{equation}
\label{nominal and residual}
\mathbf{x}_{k+1} = \mathbf{f}(\mathbf{x}_k,\mathbf{u}_k)+\mathbf{B}_d\mathbf{g}(\mathbf{z}_k),
\end{equation}
where $\mathbf{f}$ represents the nominal model and $\mathbf{g}$ is the residual model. The residual model is assumed to only affect the subspace spanned by $\mathbf{B}_d$. Feature vector $\mathbf{z}_k$ of residual model is extracted from $\mathbf{x}_k$ and $\mathbf{u}_k$.

\subsection{Nominal Vehicle Model}
\label{nominal vehicle model}
This paper chooses a single-track dynamics model with nonlinear tire forces as the nominal model to extract the essential vehicle dynamics, as formulated below:
\begin{equation}
\label{nominal model equation}
\dot{\mathbf{x}}
= \begin{bmatrix} 
v_x\mathrm{cos}\varphi-v_y\mathrm{sin}\varphi \\ 
v_x\mathrm{sin}\varphi+v_y\mathrm{cos}\varphi\\ 
\omega \\
\frac{1}{m}(F_{r,x}-F_d-F_{f,y}\mathrm{sin}\delta+F_{f,x}\mathrm{cos}\delta)+v_{y}\omega \\
\frac{1}{m}(F_{r,y}+F_{f,y}\mathrm{cos}\delta+F_{f,x}\mathrm{sin}\delta)-v_{x}\omega \\
\frac{1}{I_{z}}((F_{f,y}\mathrm{cos}\delta+F_{f,x}\mathrm{sin}\delta)l_f-F_{r,y}l_r) \\
\Delta T \\
\Delta \delta 
\end{bmatrix}.
\end{equation}
Here, $F_{f,x}$ and $F_{r,x}$ denote the longitudinal force on the front and rear wheels, respectively, while $F_{f,y}$ and $F_{r,y}$ denote the lateral force on the front and rear wheels. $F_d$ represents the air drag. $m$ is the vehicle mass and $I_z$ is the yaw moment of inertia. $l_f$ and $l_r$ denote the distances between the center of gravity (CoG) and the front/rear axle, respectively. 
In the equations, $F_{f,y}$ and $F_{r,y}$ are calculated using a simplified magic formula \cite{pacejka2005tire}, based on the values of the front and rear wheel side slip angles $\alpha_f$ and $\alpha_r$:
\begin{equation}
F_{f/r,y} = D_{f/r}\mathrm{sin}(C_{f/r}\mathrm{arctan}(B_{f/r}\alpha_{f/r})).
\label{magic formula}
\end{equation}
The parameters $B_{f/r},C_{f/r},D_{f/r}$ are tire parameters used in the Magic Formula. The front and rear wheel side slip angles are calculated based on kinematic relationships:
\begin{equation}
\begin{aligned}
\label{side slip angle}
\alpha_f = \delta-\mathrm{arctan}(\frac{v_y+l_f\omega}{v_x})
\\
\alpha_r = \mathrm{arctan}(\frac{-v_y+l_r\omega}{v_x}).
\end{aligned}
\end{equation}
The longitudinal tire forces are calculated based on the command torque: 
\begin{equation}
\label{driving force}
F_{f,x} = \kappa \frac{T}{r}-C_{f,r},F_{r,x} = (1-\kappa)\frac{T}{r}-C_{r,r},
\end{equation}
where the torque distribution coefficient for the front axle is $\kappa$ and $r$ represents the rolling radius of the tire. $C_{f,r}$ and $C_{r,r}$ denote the rolling resistance for the front and rear tires. The air drag $F_d$ is modeled as:
\begin{equation}
\label{equation: air drag}
F_{d} = C_w v_x^2,
\end{equation}
where $C_w$ denotes the coefficient of air drag. When applied to MPC, the nominal model in (\ref{nominal model equation}) is linearized and discretized at each time step along the prediction horizon with associated operation points, resulting in the discrete nominal model $\mathbf{f}(\mathbf{x}_k,\mathbf{u}_k)$ as in (\ref{nominal and residual}).

\subsection{Analysis of Nominal Model Discrepancies}
\label{Analysis of Nominal Model Discrepancies}
While the previously presented nominal model captures the essential vehicle dynamics, discrepancies remain between the model and actual vehicle dynamics.
In the following, the sources of discrepancies in the nominal model are analyzed to inform the design of the residual model. 
Based on the structure of the nominal model, the dynamics of the first three states, \(X\), \(Y\), and \(\varphi\), are fully governed by kinematic relationships, while the last two states, \(T\) and \(\delta\), represent actuator outputs. 
The dynamics of these states are assumed to be sufficiently accurate, with model errors only affecting the velocity states \(v_x\), \(v_y\), and \(\omega\). 
Consequently, the matrix \(\mathbf{B}_d\) is defined as \(\mathbf{B}_d=\begin{bmatrix} \mathbf{0}_{3\times3} & \mathbf{I}_{3\times3} & \mathbf{0}_{3\times2} \end{bmatrix}^T\). 
The dynamics model corresponding to the velocity states can be expressed in matrix form as follows:
\begin{equation}
\label{dynamic state equation}
\begin{aligned}
\begin{bmatrix} \dot{v}_x \\ \dot{v}_y \\ \dot{\omega} \end{bmatrix} = & \begin{bmatrix} 
\cos\delta/m & -\sin\delta/m & 1/m & 0\\ \sin\delta/m & \cos\delta/m & 0 & 1/m \\ l_f\sin\delta/I_z & l_r\cos\delta/I_z & 0 & -l_r/I_z\end{bmatrix}\begin{bmatrix} F_{f,x} \\ F_{f,y} \\ F_{r,x} \\ F_{r,y} \end{bmatrix} \\ &+\begin{bmatrix} v_y\omega - C_wv_x^2 \\ -v_x\omega \\ 0 \end{bmatrix}.
\end{aligned}
\end{equation}

In nominal vehicle model, parameters $m$, $l_f$, $l_r$, $I_z$ and $C_w$ are considered invariable elements as they can be precisely measured through experimental calibration and exhibit negligible variation during operation. Conversely, longitudinal and lateral tire forces $[F_{f,x},F_{f,y},F_{r,x},F_{r,y}]^T$ are considered the variable elements. The tire forces may deviate from the model during operation due to factors such as road conditions, tire pressure, temperature and tire wear. Additionally, the coupling between longitudinal and lateral tire forces further impact model accuracy. These deviations in tire forces constitute the primary source of discrepancies in the nominal model. Consequently, the model error corresponding to the velocity states $[\Delta\dot{v}_x,\Delta\dot{v}_y,\Delta\dot{\omega}]^T$ can be expressed as follows:
\setlength{\arraycolsep}{3pt} 
\begin{equation}
\label{error dynamics equation}
\begin{bmatrix} \Delta\dot{v}_x \\ \Delta\dot{v}_y \\ \Delta\dot{\omega} \end{bmatrix} = \begin{bmatrix} \cos\delta/m & -\sin\delta/m & 1/m & 0\\ \sin\delta/m & \cos\delta/m & 0 & 1/m \\ l_f\sin\delta/I_z & l_r\cos\delta/I_z & 0 & -l_r/I_z\end{bmatrix}\begin{bmatrix} \Delta F_{f,x} \\ \Delta F_{f,y} \\ \Delta F_{r,x} \\ \Delta F_{r,y} \end{bmatrix}
\end{equation}
where $[\Delta F_{f,x},\Delta F_{f,y},\Delta F_{r,x},\Delta F_{r,y}]^T$ represent the dieviation of tire forces. Considering that the front wheel steering angle $\delta$ is generally small, and the deviations in tire forces $\Delta F_{x/y,f/r}$ are also small relative to the values of tire model $F_{x/y,f/r}$, the impact of higher-order terms $\Delta F_{f/r,x/y} \sin\delta$ can be neglected, and $\cos\delta \approx 1$. Consequently, the expression for model error can be simplified as follows:
\begin{equation}
\label{final error dynamics equation}
\begin{bmatrix} \Delta\dot{v}_x \\ \Delta\dot{v}_y \\ \Delta\dot{\omega} \end{bmatrix} \approx \begin{bmatrix} (\Delta F_{r,x}+\Delta F_{f,x})/m \\ (\Delta F_{r,y}+\Delta F_{f,y})/m \\ (\Delta F_{f,y}l_f-\Delta F_{r,y}l_r)/I_z \end{bmatrix}
\end{equation}

\subsection{Low Dimensional Residual Model}
\label{low dimensional residual model}
To incrementally refine the accuracy of the nominal model and adapt to vehicle dynamics variations, GPs are employed in the formulation of residual model, leveraging their flexibility and probabilistic nature to learn deviations caused by the variable elements.
The training set required for GPs is constructed from real-time collected data. Training data $\mathbf{y}_j$ is derived by calculating the deviation between the sensor measurements and the prediction provided by the nominal model:
\begin{equation}
\label{deviations calculation}
\mathbf{y}_k = \mathbf{g}(\mathbf{z}_k)+\mathbf{w}_k = \mathbf{B}_d^{\dagger}(\mathbf{x}_{k+1}-\mathbf{f}(\mathbf{x}_k,\mathbf{u}_k)),
\end{equation}
where $\mathbf{w}_j \sim \mathcal{N}(\mathbf{0},\mathbf{\Sigma}_w)$ is gaussian noise with variance $\mathbf{\Sigma}_w $= diag$([\sigma_1^2,\sigma_2^2,\sigma_3^2])$. $\mathbf{B}_d^{\dagger}$ is the Moore-Penrose pseudo-inverse of $\mathbf{B}_d$. To enhance the coverage of feature space by the training set, a data selection procedure based on the independence metric proposed in \cite{nguyen2011incremental} is applied to continuously maintain a training set of size $m$:
\begin{equation}
\label{dataset}
\cal{D} = \begin{bmatrix}
    \mathbf{Z} = {\begin{bmatrix}
        \mathbf{z}_0^T; \cdots ;\mathbf{z}_m^T
    \end{bmatrix}},
    \mathbf{Y} = {\begin{bmatrix}
        \mathbf{y}_0^T; \cdots ;\mathbf{y}_m^T
    \end{bmatrix}}
\end{bmatrix}.
\end{equation}

Furthermore, each dimension of the residual model output is assumed to be uncorrelated with the others. Under this assumption, GPs can provide posterior predictions of the mean $\mu_a(\mathbf{z}_k)$ and variance $\Sigma_a(\mathbf{z}_k)$ of the nominal model error in dimension $a$ at test point $\mathbf{z}_*$:
\begin{equation}
\begin{aligned}
    \mu_a(\mathbf{z}_*) &= \mathbf{k}_*^T (\mathbf{K} + \sigma_a^2 \mathbf{I})^{-1} [\mathbf{Y}]_{*,a}, \\
    \Sigma_a(\mathbf{z}_*) &= k(\mathbf{z}_*, \mathbf{z}_*) - \mathbf{k}_*^T (\mathbf{K} + \sigma_a^2 \mathbf{I})^{-1} \mathbf{k}_*.
\end{aligned}
\end{equation}
The matrix $\mathbf{K}$ denotes the covariance evaluated at all pairs of points in training set and $\begin{bmatrix} \mathbf{K} \end{bmatrix}_{ij} = k(\mathbf{z}_i,\mathbf{z}_j)$. The vector $\mathbf{k}_* = \begin{bmatrix} k(\mathbf{z}_*,\mathbf{z}_1), \cdots ,k(\mathbf{z}_*,\mathbf{z}_m) \end{bmatrix}^T$. The Squared Exponential (SE) kernel function is employed to define the GP kernel $k(\mathbf{z}_i,\mathbf{z}_j)$:
\begin{equation}
\label{SE kernel function}
k(\mathbf{z}_i,\mathbf{z}_j) = \sigma_f^2 \mathrm{exp}(-\frac{1}{2}(\mathbf{z}_i-\mathbf{z}_j)^T\mathbf{M}(\mathbf{z}_i-\mathbf{z}_j)),
\end{equation}
where parameter $\mathbf{M}$ defining the length-scale matrix and parameter $\sigma_f^2$ defining the squared signal variance.

The final multivariate GPs approximation of the unknown residual $\mathbf{g}(\mathbf{z}_*)$ is obtained by aggregating the outputs for each dimension:
 \begin{equation}
     \mathbf{g}(\mathbf{z}_*) \sim \mathcal{N}(\boldsymbol{\mu},\mathbf{\Sigma}),
 \end{equation}
 where $\boldsymbol{\mu} = \begin{bmatrix} \mu_{v_x},\mu_{v_y},\mu_{\omega} \end{bmatrix}^T$ and $\mathbf{\Sigma} = $diag$([\Sigma_{v_x},\Sigma_{v_y},\Sigma_{\omega}]^T)$.

The features for GP are critical. In previous researches, the features for residual model corresponding to the single-track dynamic model are typically selected as the dynamic part of the vehicle states $\begin{bmatrix} v_x, v_y, \omega, \delta, T \end{bmatrix} $, spanning a 5-dimensional feature space. This necessitates a vast training set to cover the entire feature space, leading to unacceptable computational burdens. Additionally, it is infeasible to control the vehicle to traverse every region within this 5-dimensional space. As a result, previous researches have achieved satisfactory performance only under specific, fixed operating conditions. Accordingly, this paper extract physical variables most correlated with nominal model errors as features, aiming to reduce the dimensionality of the feature space. 

As shown in (\ref{final error dynamics equation}), the primary source of nominal model error arises from discrepancies between actual tire forces and the tire model. The lateral tire forces in nominal model are functions of the tire side slip angles; therefore, deviations in lateral tire forces, $\Delta F_{f/r,y}$, are primarily functions of the tire side slip angle $\alpha_{f/r}$. 
Similarly, deviations in longitudinal tire force, $\Delta F_{f/r,x}$, are primarily due to differences between the tire force predicted based on the commanded torque and actual tire force, making $\Delta F_{f/r,x}$ primarily a function of the command torque $T$. 
When accounting for coupling effects between lateral and longitudinal tire forces, tire force deviations $\Delta F_{f/r,x,y}$ depend on both the command torque $T$ and the tire slip angle $\alpha_{f/r}$.
Consequently, the residual of the vehicle dynamics model can be formulated as a function of the front wheel side slip angle $\alpha_f$, rear wheel side slip angle $\alpha_r$, and command torque $T$. Thus, these three variables are selected as the features for the residual model:

\begin{align}
\label{residual model input features}
\mathbf{z} = \begin{bmatrix} \alpha_f, \alpha_r, T\end{bmatrix}^T.
\end{align}

Based on the preceding analysis, the dimensionality of the features for the residual model is reduced from 5 to 3. This reduction in dimensionality significantly decreases the size of training set required to cover the entire feature space.

\subsection{Constrains for Features}
\label{subsection: Constraints for features}
In (\ref{residual model input features}), the 3-dimensional features space is defined. To explicitly define the valid region in the feature space and reduce the region that the training set needs to cover, three physical constraints related to the features are designed. These constraints eliminate invalid regions, and is subsequently utilized in the MPC controller. Data violating these constraints are considered invalid and excluded. 

The first constraint concerns tire force limits. To ensure vehicle safety during operation, the total tire forces should not exceed the friction limits to prevent instability and hazardous conditions. Therefore, the combined longitudinal and lateral forces of the front and rear tires are constrained within the tire-road contact ellipse:
\begin{equation}
\label{constrain1}
    (p_{long}F_{f/r,x})^2+F_{f/r,y}^2 \leq (p_{ellipse}D_{f/r})^2,
\end{equation}
with $p_{long}$ and $p_{ellipse}$ defining the ellipse shape.
The second constraint concerns the tire slip angle limits. By restricting the maximum allowable slip angles for tires, this constraint seeks to prevent the tires from reaching the lateral force saturation zone:
\begin{equation}
\label{constrain2}
    -\alpha_{max} \leq \alpha_{f/r} \leq \alpha_{max}.
\end{equation}

\begin{figure}[!t]
\centering
\includegraphics[width=0.9\linewidth]{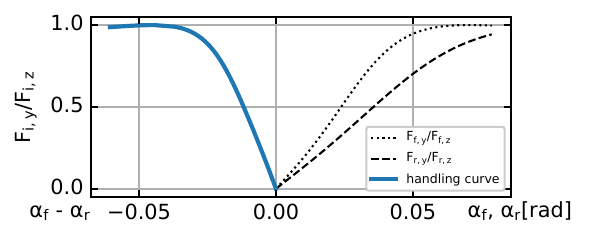}
\vspace{-2.0ex}
\caption{Handling diagram of vehicle.}
\label{handling diagram}
\vspace{-4.0ex}
\end{figure}

The third constraint is derived from the vehicle handling diagram \cite{rossa2012bifurcation}, as shown in Fig. \ref{handling diagram}. The curves represent the relationship between the effective axle characteristics \(F_{f/r,y}/ F_{f/r,z}\) and the tire slip angles under steady-state steering. Analysis of the handling curves indicates that, beyond a certain threshold, further increases in the difference \(\alpha_f - \alpha_r\) cease to enhance the lateral acceleration of vehicle. Consequently, the difference \(\alpha_f - \alpha_r\) should be constrained within a reasonable range:

\begin{equation}
\label{constrain3}
-\Delta \alpha_{max} \leq \alpha_f - \alpha_r \leq \Delta \alpha_{max}.
\end{equation}

\begin{figure}[!t]
\centering
\vspace{1.0ex}
\includegraphics[width=0.6\linewidth, trim=0 0 0 40,clip]{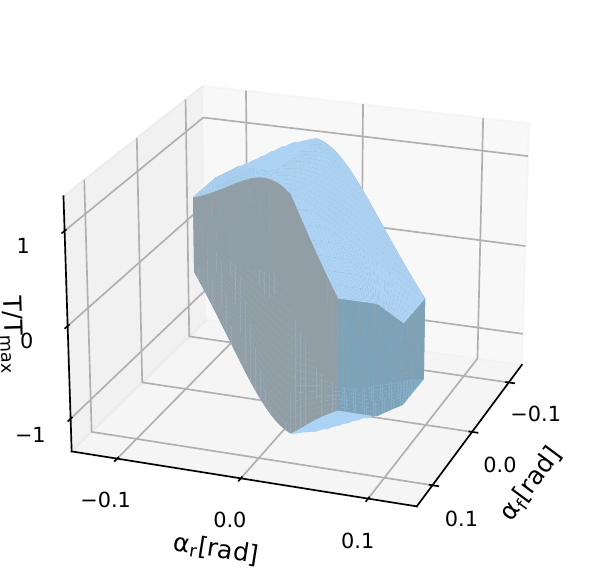}
\vspace{-2.0ex}
\caption{Valid region within the feature space.}
\label{valid region}
\vspace{-3.0ex}
\end{figure}

Building on the three aforementioned constraints, a finite valid region within the feature space is established. 
This region encompasses the majority of the dynamic states encountered during operation, explicitly define the scope within feature space that the training set must cover. The constructed valid region is depicted in the Fig. \ref{valid region}.

\section{LEARNING BASED MPC FORMULATION}
\label{learning based mpc formulation}

In this section, the proposed vehicle model is integrated into a MPC controller. To fully exploit the benefits of the residual model, we focus on the challenging scenario of autonomous racing which is formulated as a Model Predictive Contouring Control (MPCC) problem \cite{liniger2015optimization}.
\subsection{Contouring Control}

In MPCC, the trajectory planning and tracking tasks are formulated as a unified optimization problem that determines the control inputs required for the race car to follow the racing track. The track is parameterized by $\theta \in [0, \theta_{\text{max}}]$, where $\theta$ designates a parameter that specifies the corresponding centerline position and orientation, denoted as $[X_c(\theta), Y_c(\theta), \Phi_c(\theta)]$. The state variables in (\ref{nominal model equation}) are augmented with $\theta$ and $v_s$, which approximate the vehicle progression along the track and its velocity relative to the centerline. These parameters adhere to the update equation $\theta_{k+1} = \theta_k + v_s \Delta t$. The positional deviations between the vehicle position and the corresponding reference point on the track centerline are decomposed into the lag error $e_l$ and contour error $e_c$, defined as follows:
\begin{equation}
\label{MPC formulation}
\begin{aligned}
e_l(\mathbf{x}_k,\theta_k)= &-\mathrm{cos}(\Phi(\theta_k))(X_k-X_c(\theta_k)) \\
&-\mathrm{sin}(\Phi(\theta_k))(Y_k-Y_c(\theta_k)), \\
e_c(\mathbf{x}_k,\theta_k)= &\mathrm{sin}(\Phi(\theta_k))(X_k-X_c(\theta_k)) \\
&-\mathrm{cos}(\Phi(\theta_k))(Y_k-Y_c(\theta_k)).
\end{aligned}
\end{equation}

\subsection{MPCC Formulation}
The formulation of MPCC is given by:
\begin{equation}
\label{MPC formulation}
\begin{aligned}
\min_{\{\mathbf{u}_k,v_k\}} \sum_{k=1}^NJ(\mathbf{x}_k,\theta_k,v_k)+U(\mathbf{x}_k,\mathbf{u}_k)-q_v v_k 
\\ s.t.
\begin{array}{l}
\mathbf{x}_{k+1} = \mathbf{f}(\mathbf{x}_k,\mathbf{u}_k)+\mathbf{B}_d\mathbf{g}(\mathbf{z}_k), \\
\theta_{k+1}=\theta_k+v_s \Delta t,\\
\|X_k - X_c(\theta_k)\|^2 + \|X_k - X_c(\theta_k)\|^2 \leq R^2, \\
\mathbf{x}_{min} \leq \mathbf{x}_{k} \leq \mathbf{x}_{max}, \mathbf{u}_{min} \leq \mathbf{u}_{k} \leq \mathbf{u}_{max},\\
(\ref{constrain1}), (\ref{constrain2}), (\ref{constrain3}).
\end{array}
\end{aligned}
\end{equation}

The cost function in (\ref{MPC formulation}) is composed of three primary components. The first term, $J(\mathbf{x}_k,\theta_k,v_k) = q_l e_l(\mathbf{x}_k,\theta_k)^2 + q_c e_c(\mathbf{x}_k,\theta_k)^2$, penalizes the contouring error. The regularization term  $U(\mathbf{x}_k,\mathbf{u}_k) = \|\mathbf{x}_k\|_{\mathbf{R}_\mathbf{x}}^2 + \|\mathbf{u}_k\|_{\mathbf{R}_\mathbf{u}}^2$ imposes a penalty on the control inputs and their changing rates, ensuring that the control actions are smooth and preventing abrupt changes in trajectory. The third term, $q_v v_k$, is designed to maximize the vehicle progression along the track within the prediction horizon. Here, $q_l$, $q_c$, and $q_v$ are constant weights, and $\mathbf{R}_{\mathbf{x}}$ and $\mathbf{R}_{\mathbf{u}}$ are weighting matrices.

The constraints outlined in (\ref{MPC formulation}) impose multiple limitations. Beyond ensuring consistency with the vehicle model in (\ref{nominal and residual}), the track constraints confine the vehicle position within the track radius $R$, while the state and control constraints ensure that vehicle states and control inputs adhere to the physical limitations. The constraints in Section \ref{subsection: Constraints for features} restrict the vehicle state within the valid region, preventing unsafe behaviors. To guarantee the feasibility of the optimization problem, all constraints are modeled as soft constraints.

\section{RESULTS}
\label{results}

\begin{figure}[!t]
\centering
\includegraphics[width=\columnwidth]{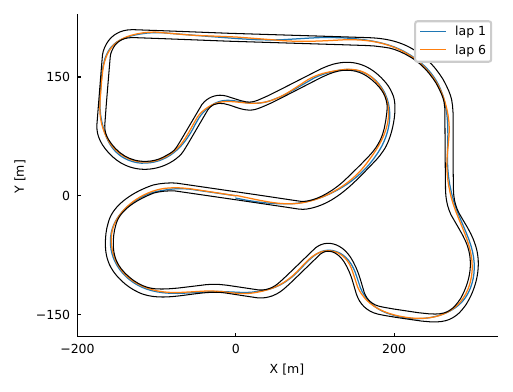}
\vspace{-5.0ex}
\caption{Racing trajectory of lap 1 and lap 6.}
\label{simulation racing trajectory}
\setlength{\abovecaptionskip}{0.cm}
\vspace{-1.0ex}
\end{figure}

\begin{figure}[!t]
\vspace{-2.0ex}
\centering
\begin{subfigure}{0.49\linewidth}
    \centering
        \setlength{\abovecaptionskip}{0.cm}
    \includegraphics[width=1.0682\linewidth]{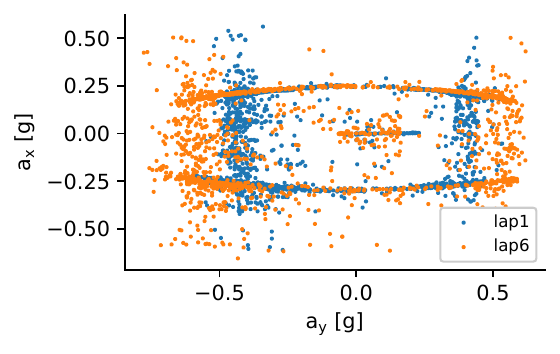}
    \vspace{-3.0ex}
    \caption{GG-diagram}
    \label{simulation comparison gg diagram}
\end{subfigure}
\centering
\begin{subfigure}{0.49\linewidth}
    \centering
        \setlength{\abovecaptionskip}{0.cm}
    \includegraphics[width=1.0\linewidth]{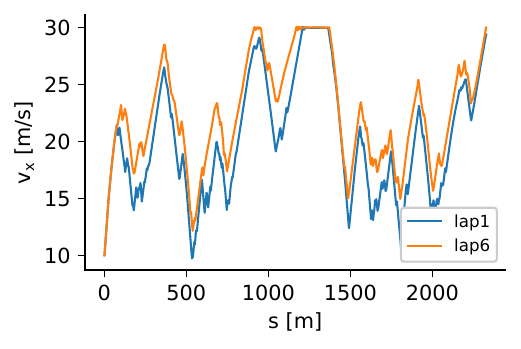}
    \vspace{-3.0ex}
    \caption{Velocity profiles}
    \label{simulation comparison vx-s}
\end{subfigure}
\caption{Comparison of nominal (lap 1) and learning based MPC controller (lap 6).}
\label{simulation comparison}
\vspace{-1.0ex}
\end{figure}

\begin{figure}[!t]
\vspace{-1.0em}
\centering
\setlength{\abovecaptionskip}{0.cm}
\begin{subfigure}{0.49\linewidth}
    \centering
        \includegraphics[width=1.05\linewidth]{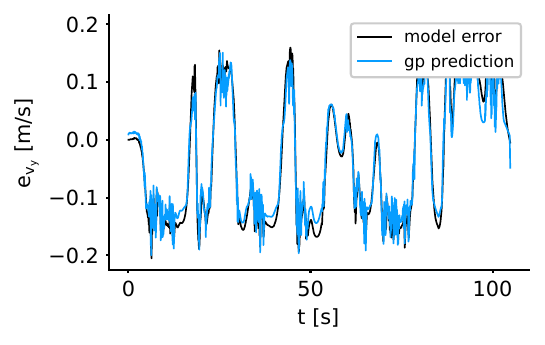}
        \vspace{-4.0ex}
    \caption{Error in $v_y$}
    \label{carsim_error_vy}
\end{subfigure}
\centering
\begin{subfigure}{0.49\linewidth}
    \centering
        \includegraphics[width=1.05\linewidth]{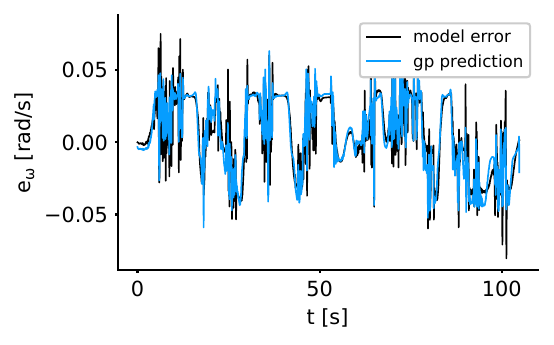}
        \vspace{-4.0ex}
    \caption{Error in $\omega$}
    \label{carsim_error_omega}
\end{subfigure}
\caption{Model error and GP prediction in lap 6}
\label{simulation error}
\vspace{-2.0ex}
\end{figure}

\begin{table}[!t]
\caption{\label{tab:lap} \textbf{Experimental Results in Simulation}}
\label{experimental results in simulation}
\centering
\begin{tabular}{ccccc}
\toprule
Lap & Time (s) & $\Vert a_y \Vert_{\text{max}}$ (g) & Data Updates & Average Speed (m/s)\\
\midrule
1&119.63&0.59&-&19.46 \\
2&111.20&0.68&82&20.94 \\
3&107.60&0.74&78&21.64 \\
4&105.43&0.71&20&22.08 \\
5&104.93&0.72&6&22.18 \\
6&104.85&0.73&9&22.20 \\
\bottomrule
\end{tabular}
\vspace{-5.0ex}
\end{table}


\begin{table}[t]
\caption{\label{tab:error} \textbf{Model Error in Simulation}}
\label{model prediction error in simulation}
\centering
\begin{tabular}{ccccc}
\toprule
\multirow{2}{*}{Lap} & \multicolumn{2}{c}{$e_{v_y}$ ($10^{-2}$m/s)} & \multicolumn{2}{c}{$e_{\omega}$ ($10^{-2}$rad/s)}\\
& nominal & nominal & nominal & nominal\\
& model & +residual & model & +residual \\
\midrule
1 & 7.26$\pm$2.87 & - & 1.87$\pm$1.16 & - \\
2 & 8.99$\pm$4.30 & 1.83$\pm$1.92 & 1.88$\pm$1.35 & 0.68$\pm$0.65 \\
3 & 10.02$\pm$4.75 & 1.73$\pm$2.51 & 2.09$\pm$1.47 & 0.78$\pm$0.81 \\
4 & 10.61$\pm$4.67 & 1.61$\pm$1.62 & 2.04$\pm$1.40 & 0.67$\pm$0.71 \\
5 & 10.62$\pm$4.82 & 1.53$\pm$1.34 & 2.03$\pm$1.43 & 0.64$\pm$0.66 \\
6 & 10.64$\pm$4.87 & 1.67$\pm$1.28 & 2.03$\pm$1.40 & 0.66$\pm$0.64 \\
\bottomrule
\end{tabular}
\vspace{-2.0ex}
\end{table}

In this section, the effectiveness of the proposed controller is validated through autonomous racing tasks in both simulation and actual vehicles. The specific algorithm implementation and experimental details are presented in the following.
\subsection{Controller Implementation}
The controller is deployed on a Nuvo-9160 industrial computer with i9-13900E CPU, operates with a sampling time of \(\Delta t = 50 \, \text{ms}\) and a prediction horizon of \(N = 80\). Optimization is performed using the HPIPM library \cite{frison2020hpipm}, with a maximum of 40 iterations to constrain the computation time. 
Control signals and state information of vehicle are transmitted via TCP/UDP communication protocols. 

\subsection{Simulation Experiment}

The proposed controller is first validated within Carsim simulation environment. The simulation vehicle is chosen as B-Class Hatchback 4-wheel and the handling course race line, with a total length of 2327.98m, is selected as the racing track. The track has a width of 10m and the friction coefficient is 1.0. The invariable elements of the nominal vehicle model are provided by CarSim. In the experiment, the initial speed is 10 m/s, with a maximum speed limit of 30 m/s. The vehicle completes the first lap with nominal MPC controller as a baseline. The collected data is filtered using the data selection mechanism in \cite{nguyen2011incremental} to construct a training set of size 100, serving as the initial training set for residual model. A total of 5 laps are conducted using the proposed learning based MPC controller, with data collected after each lap used for offline updates of training set for the subsequent laps. The vehicle trajectories from the first and final laps are shown in Fig. \(\ref{simulation racing trajectory}\).

To quantify the performance of proposed controller, the key metrics are summarized in Table \ref{experimental results in simulation}. The proposed residual model consistently improves lap times and average speed. The initial lap time of 119.63s, obtained using the nominal controller, is reduced to 104.85s after five laps of data updates, representing an overall improvement of 12.35\%. The sustained increase in maximum lateral acceleration during racing demonstrates the enhanced utilization of tire dynamics, which is also illustrated in Fig. \ref{simulation comparison}(\subref{simulation comparison gg diagram}). Fig. \ref{simulation comparison}(\subref{simulation comparison vx-s}) compares the vehicle speeds at each track position between the first and final laps, showing significant speed improvements at every location with the learning based MPC controller.

The predictive performance of the model is also evaluated. At each time step, the deviation between the actual system states and the model predictions is quantified using \(\Vert \mathbf{e} \Vert_{\text{nom}} = \Vert \mathbf{x}_{k+1} - \mathbf{f}(\mathbf{x}_k, \mathbf{u}_k) \Vert\) and \(\Vert \mathbf{e} \Vert_{\text{GP}} = \Vert \mathbf{x}_{k+1} - \mathbf{f}(\mathbf{x}_k, \mathbf{u}_k) - \mathbf{B}_d \mathbf{g}(\mathbf{z}_k) \Vert\). Table \ref{model prediction error in simulation} outlines the mean and standard deviation of prediction errors \(e_{v_y},\) and \(e_{\omega}\) during racing, affirming that the proposed residual model substantially mitigates the prediction discrepancies of the nominal model. Fig. \ref{simulation error} provides a graphical comparison of the actual error \(e_{v_y}\) and \(e_{\omega}\), and the GP predictions during the 6th lap, highlighting the superior predictive capability of the residual model. 

\subsection{Hardware Experiment}

\begin{figure*}[!t]
\centering
\subfloat[\label{hardware comparison v-s} Velocity profiles]{
		\includegraphics[scale=0.90]{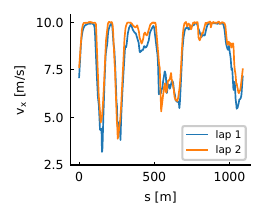}
  \hspace{-1.5em}
  \vspace{-5.0ex}
  }
\subfloat[\label{hardware comparison trajectory}Racing trajectory of lap 1 and lap 2]{
		\includegraphics[scale=0.90]{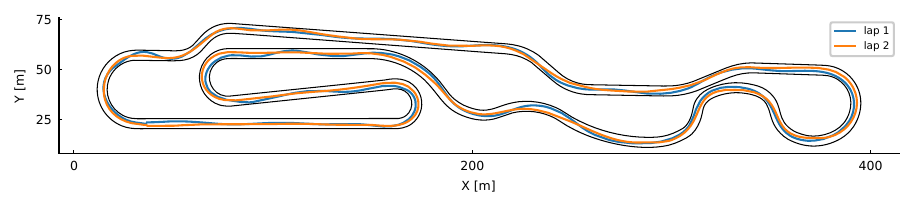}
  \hspace{-1.5em}
  \vspace{-5.0ex}
  }
\vspace{-1.0ex}
\caption{Comparison of nominal (lap 1) and learning based MPC controller (lap 2).}
\label{hardware comparison}
\setlength{\abovecaptionskip}{0.cm}
\vspace{-3.0ex}
\end{figure*}

\begin{figure}[!t]
\centering
\includegraphics[width=0.70\linewidth]{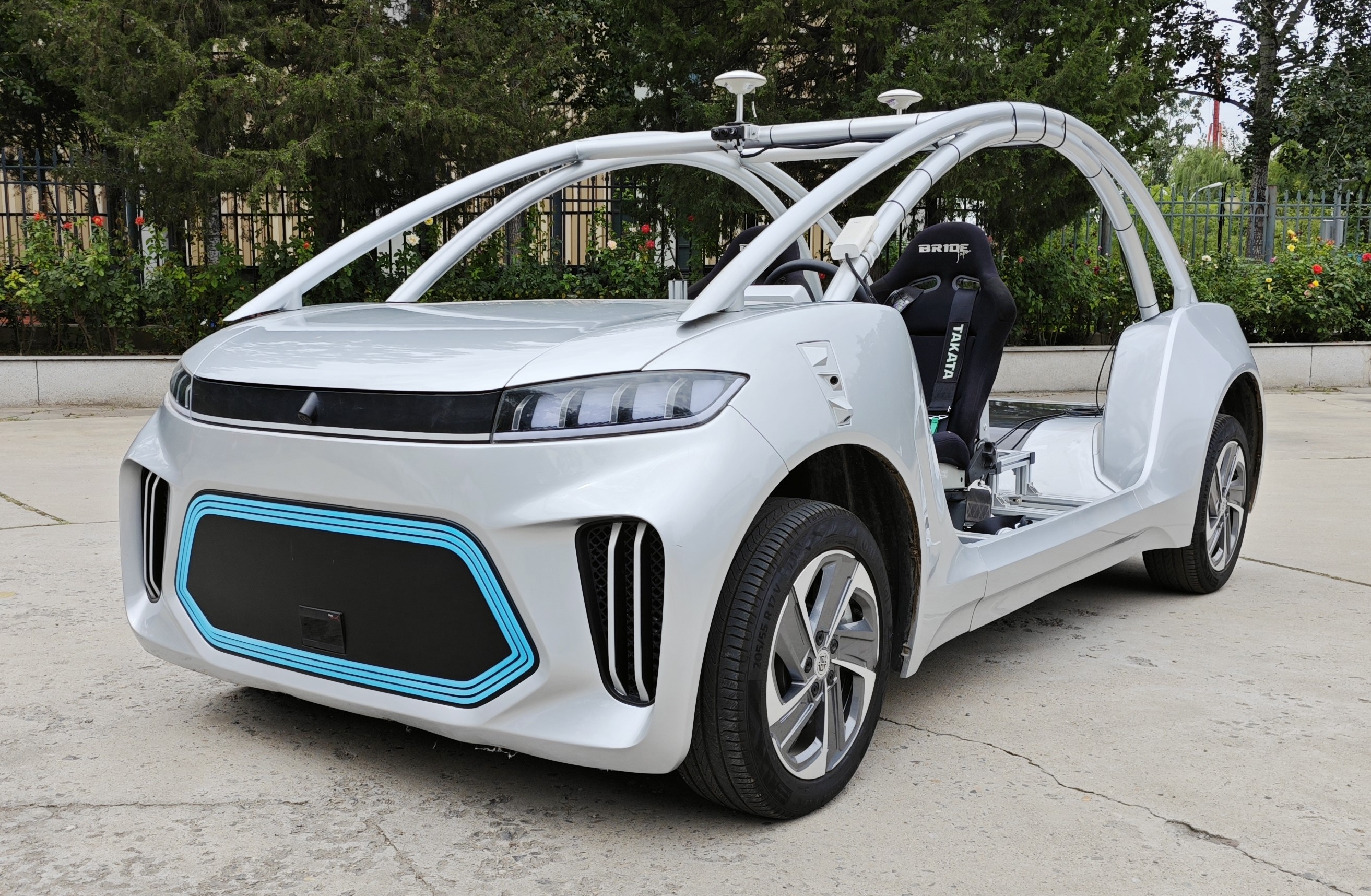}
\caption{The real vehicle testing platform.}
\label{violetlight}
\vspace{-4.0ex}
\end{figure}

The proposed controller is further validated on a real vehicle testing platform as shown in Fig \ref{violetlight}. The vehicle states are measured using the onboard RTK. The racing track including multiple straights and curves is 1087m long, 10m wide, with a concrete surface. The invariable elements of the nominal model are derived from comprehensive calibration experiments. To ensure safety, speed is limited to 10m/s, and the maximum tire forces are restricted to 50\% of the road adhesion limit. Both the nominal and the learning based MPC controller are each employed to complete one lap of racing. The training set of GP is constructed by filtering data in lap 1, consisting of 100 points.

\begin{table}[!t]
\caption{\label{tab:lap} \textbf{Hardware Experimental Results}}
\label{hardware experimental results}
\centering
\begin{tabular}{ccccc}
\toprule
Lap & Time (s) & $\Vert a_y \Vert_{\text{max}}$ (g) & Average Speed (m/s)\\
\midrule
1&136.38&0.45&7.97 \\
2&131.40&0.47&8.27 \\
\bottomrule
\end{tabular}
\end{table}

\begin{figure}[!t]
\centering
\setlength{\abovecaptionskip}{0.cm}
\begin{subfigure}{0.49\linewidth}
\centering
    \setlength{\abovecaptionskip}{0.cm}
\includegraphics[width=1.05\linewidth]{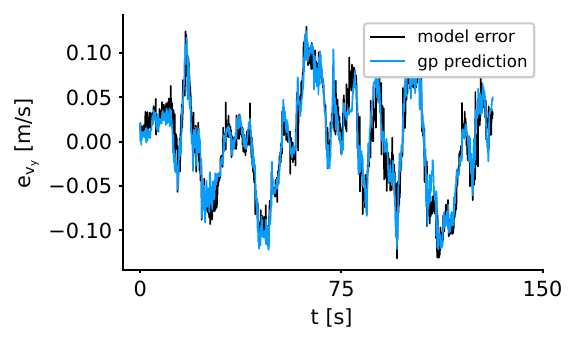}
\caption{Error in $v_y$}
\end{subfigure}
\centering
\begin{subfigure}{0.49\linewidth}
\centering
    \setlength{\abovecaptionskip}{0.cm}
\includegraphics[width=1.05\linewidth]{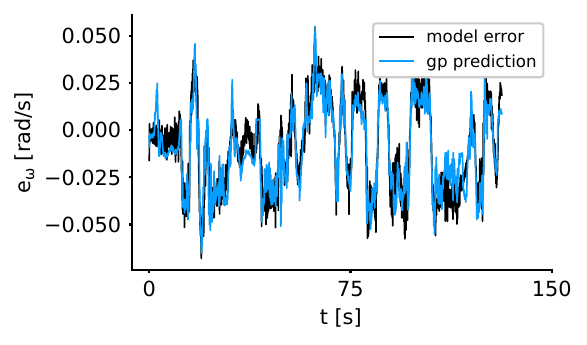}
\caption{Error in $\omega$}
\end{subfigure}
\caption{Model error and GP prediction in lap 2}
\label{hardware model error compare}
\end{figure}


\begin{table}[!t]
\caption{\label{hardware model error} \textbf{Model Error in Hardware Experiment}}
\centering
\begin{tabular}{ccccc}
\toprule
\multirow{2}{*}{Lap} & \multicolumn{2}{c}{$e_{v_y}$ ($10^{-2}$m/s)} & \multicolumn{2}{c}{$e_{\omega}$ ($10^{-2}$rad/s)}\\
& nominal & nominal & nominal & nominal\\
& model & +residual & model & +residual \\
\midrule
1 & 4.21$\pm$3.18 & - & 1.88$\pm$1.33 & - \\
2 & 4.40$\pm$3.08 & 1.57$\pm$1.46 & 1.88$\pm$1.39 & 0.87$\pm$0.72 \\
\bottomrule
\end{tabular}
\end{table}

\begin{figure}[!t]
\centering
\setlength{\abovecaptionskip}{0.cm}
\begin{subfigure}{0.49\linewidth}
    \centering
        \includegraphics[width=1.05\linewidth, trim=0 0 0 40,clip]{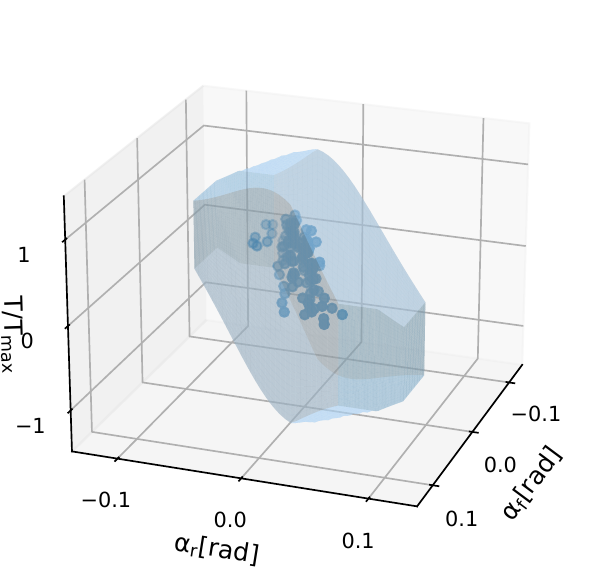}
    \caption{Data in training set}
    \label{lap2_dictionary}
\end{subfigure}
\centering
\begin{subfigure}{0.49\linewidth}
    \centering
        \includegraphics[width=1.05\linewidth, trim=0 0 0 40,clip]{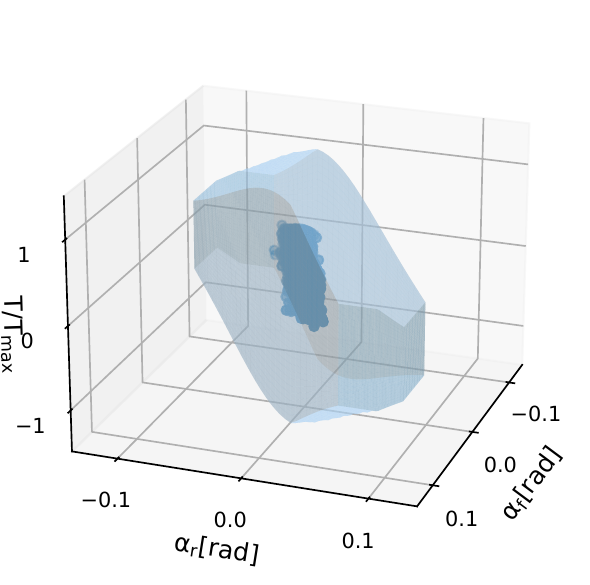}
    \caption{Data in lap2}
    \label{lap2_data}
\end{subfigure}
\caption{Data distribution within the feature space}
\label{data distribution within the feature space}
\end{figure}

The key performance metrics from both laps are summarized in Table \ref{hardware experimental results}, indicating a 5s improvement in lap time achieved by the proposed controller. Fig. \ref{hardware comparison} provides a comparison of the longitudinal velocity at each point on the track and the trajectories from 2 laps. Fig. \ref{hardware model error compare} compares the residual model prediction \(e_{v_y}\) and \(e_{\omega}\) of the nominal model errors and the actual deviations during lap 2, demonstrating the capability of residual model to accurately capture the discrepancies in nominal model. Table \ref{hardware model error} summarizes the mean and standard deviation of nominal model prediction errors \(e_{v_y},\) and \(e_{\omega}\) before and after the inclusion of the residual model, highlighting the improvement in alignment with the true vehicle dynamics. Figure \ref{data distribution within the feature space} illustrates the distribution of the training dataset and lap 2 data within the residual model feature space, confirming that the training set only needs to cover a constrained valid region to deliver robust performance across varied operating conditions. These results validate that the proposed controller significantly enhances both model accuracy and control performance.

\section{CONCLUSIONS}
\label{conclusions}

This paper proposes a learning based MPC controller using a low dimensional residual model for autonomous driving. The nominal vehicle model is decomposed into variable and invariable elements. The accuracy of the invariable component is ensured through experimental measurements, while the discrepancies in variable elements are learned via a low dimensional residual model. This decomposition reduces the dimensionality of the residual model features. Physical constraints among features of residual model are formulated to explicitly define the valid region within feature space. The proposed vehicle model and constraints are incorporated into the MPC and the controller is validated through high-fidelity simulations and real-world vehicle experiments. Experimental results demonstrate that the proposed controller significantly improves the accuracy of the nominal model and enhances controller performance. 




\end{document}